\documentclass[3p,twocolumn,longtitle,reversenotenum,sort&compress,lefttitle]{elsarticle}

\pdfoutput=1

\usepackage[utf8]{inputenc}
\usepackage[T1]{fontenc}
\usepackage{xcolor}

\PassOptionsToPackage{hyphens}{url}\usepackage[hidelinks]{hyperref}

\usepackage{libertine}
\usepackage{microtype}

\usepackage{enumitem,amssymb}
\usepackage{tabularx,booktabs}

\makeatletter
\let\@afterindenttrue\@afterindentfalse
\makeatother

\nopreprintlinetrue

\begin{document}
\begin{frontmatter}

\title{The NCI Imaging Data Commons as a platform for reproducible research\\ in computational pathology}

\author[aff1]{Daniela~P.~Schacherer\corref{corresp}}
\ead{daniela.schacherer@mevis.fraunhofer.de}
\author[aff2]{Markus~D.~Herrmann} 
\author[aff3]{David~A.~Clunie} 
\author[aff1]{Henning~Höfener}
\author[aff4]{William~Clifford}
\author[aff4]{William~J.R.~Longabaugh}
\author[aff5]{Steve~Pieper}
\author[aff6]{Ron~Kikinis}
\author[aff6]{Andrey~Fedorov}
\author[aff1]{André~Homeyer\corref{corresp}}
\ead{andre.homeyer@mevis.fraunhofer.de}

\address[aff1]{Fraunhofer Institute for Digital Medicine MEVIS, Bremen, Germany}
\address[aff2]{Department of Pathology, Massachusetts General Hospital and Harvard Medical School, Boston, Massachusetts, USA}
\address[aff3]{PixelMed Publishing LLC, Bangor, Pennsylvania, USA}
\address[aff4]{Institute for Systems Biology, Seattle, Washington, USA}
\address[aff5]{Isomics Inc, Cambridge, Massachusetts, USA}
\address[aff6]{Department of Radiology, Brigham and Women's Hospital and Harvard Medical School, Boston, Massachusetts, USA}

\cortext[corresp]{Corresponding author}

\begin{abstract}
\noindent\textcolor{red}{{\LARGE This is an outdated preprint. An updated version is published in the journal \textit{Computer Methods and Programs in Biomedicine}:\\ \url{https://doi.org/10.1016/j.cmpb.2023.107839}}}

\noindent\textbf{Background and Objectives}: Reproducibility is a major challenge in
developing machine learning (ML)-based solutions in computational
pathology (CompPath). The NCI Imaging Data Commons (IDC) provides
\textgreater120 cancer image collections according to the FAIR
principles and is designed to be used with cloud ML services. Here, we
explore its potential to facilitate reproducibility in CompPath
research.

\noindent\textbf{Methods}: Using the IDC, we implemented two experiments in which a
representative ML-based method for classifying lung tumor tissue was
trained and/or evaluated on different datasets. To assess
reproducibility, the experiments were run multiple times with separate
but identically configured instances of common ML services.

\noindent\textbf{Results}: The AUC values of different runs of the same experiment were
generally consistent. However, we observed small variations in AUC
values of up to 0.045, indicating a practical limit to reproducibility.

\noindent\textbf{Conclusions}: We conclude that the IDC facilitates approaching the
reproducibility limit of CompPath research (i) by enabling researchers
to reuse exactly the same datasets and (ii) by integrating with cloud ML
services so that experiments can be run in identically configured
computing environments.

\end{abstract}

\end{frontmatter}

\section{Introduction}

Computational pathology (CompPath) is a new discipline that investigates
the use of computational methods for the interpretation of heterogeneous
data in clinical and anatomical pathology to improve health care in
pathology practice. A major focus area of CompPath is the computerized
analysis of digital tissue images~\citep{louis2015}. These images show
thin sections of surgical specimens or biopsies that are stained to
highlight relevant tissue structures. To cope with the high level of
complexity and variability of tissue images, virtually all
state-of-the-art methods use sophisticated machine learning (ML)
algorithms such as Convolutional Neural Networks
(CNN)~\citep{niazi2019}.

Because CompPath is applicable in a wide variety of use cases, there has
been an explosion of research on ML-based tissue analysis
methods~\citep{echle2020, cui2021}. Many methods are intended to assist
pathologists in routine diagnostic tasks such as the recognition of
tissue patterns for disease
classification~\citep{cruzroa2017, campanella2019, coudray2018, wang2020, iizuka2020}.
Beyond that, CompPath methods have also shown promise for deriving novel
biomarkers from tissue patterns that can predict outcome, genetic
mutations, or therapy response~\citep{echle2020}.

\subsection{Reproducibility challenges}

In recent years, it has become increasingly clear that reproducing the
results of published ML studies is
challenging~\citep{fell2022, hutson2018, raff2019, gundersen2022}.
Reproducibility is commonly defined as the ability to obtain
``consistent results using the same input data, computational steps,
methods, and conditions of analysis''~\citep{nasem2019}. Difficulties
related to reproducibility prevent other researchers from verifying and
reusing published results and are a critical barrier to translating
solutions into clinical practice~\citep{haibekains2020}. In most cases,
reproducibility problems seem to stem not from a lack of scientific
rigor, but from challenges to convey all details and set-up of complex
ML methods~\citep{haibekains2020, pineau2021, raff2019}. In the
following, we provide an overview of the main challenges related to ML
reproducibility and the existing approaches to address them.

The first challenge is the specification of the analysis method itself.
ML algorithms have many variables, such as the network architecture,
hyperparameters, and performance
metrics~\citep{hartley2020, renard2020, pineau2021}. ML workflows
usually consist of multiple processing steps, e.g., data selection,
preprocessing, training, evaluation~\citep{renard2020}. Small variations
in these implementation details can have significant effects on
performance. To make all these details transparent, it is crucial to
publish the underlying source code~\citep{haibekains2020}. Workflows
should be automated as much as possible to avoid errors when performing
steps manually. Jupyter notebooks have emerged as the de facto standard
to implement and communicate ML workflows~\citep{perkel2018}. By
combining software code, intermediate results and explanatory texts into
``computational narratives''~\citep{rule2019} that can be interactively
run and validated, notebooks make it easier for researchers to reproduce
and understand the work of others~\citep{perkel2018}.

The second challenge to reproducibility is the specification and setup
of the computing environment. ML workflows require significant
computational resources including, e.g., graphics or tensor processing
units (GPUs or TPUs). In addition, they often have many dependencies on
specific software versions. Minor variations in the computing
environment can significantly affect the results~\citep{gundersen2022}.
Setting up a consistent computational environment can be very expensive
and time consuming. This challenge can be partially solved by embedding
ML workflows in virtual machines or software containers like
Docker~\citep{perkel2019}. Both include all required software
dependencies so that ML workflows can be shared and run without
additional installation effort. Cloud ML services, like Google Vertex
AI, Amazon SageMaker, or Microsoft Azure Machine Learning, provide an
even more comprehensive solution. By offering preconfigured computing
environments for ML research in combination with the required
high-performance hardware, such services can further reduce the setup
effort and enable the reproduction of computationally intensive ML
workflows even if one does not own the required hardware. They also
typically provide web-based graphical user interfaces through which
Jupyter notebooks can be run and shared directly in the cloud, making it
easy for others to reproduce, verify, and reuse ML
workflows~\citep{perkel2019}.

The third challenge related to ML reproducibility is the specification
of data and its accessibility. The performance of ML methods depends
heavily on the composition of their training, validation and test
sets~\citep{gundersen2022, maierhein2018}. For current ML studies, it is
rarely possible to reproduce this composition exactly as studies are
commonly based on specific, hand-curated datasets which are only roughly
described rather than explicitly
defined~\citep{gundersen2018, hartley2020}. Also, the datasets are often
not made publicly available~\citep{haibekains2020}, or the
criteria/identifiers used to select subsets from publicly available
datasets are missing. Stakeholders from academia and industry have
defined the Findability, Accessibility, Interoperability, and
Reusability (FAIR) principles~\citep{wilkinson2016}, a set of
requirements to facilitate discovery and reuse of data. FAIR data
provision is now considered a ``must'' to make ML studies reproducible
and the FAIR principles are adopted by more and more public data
infrastructure initiatives and scientific
journals~\citep{scheffler2022}.

Reproducing CompPath studies is particularly challenging. To reveal fine
cellular details, tissue sections are imaged at microscopic resolution,
resulting in gigapixel whole-slide images (WSI)~\citep{patel2021}. Due
to the complexity and variability of tissue images~\citep{mccann2015},
it takes many---often thousands---of example WSI to develop and test
reliable ML models. Processing and managing such large amounts of data
requires extensive computing power, storage resources, and network
bandwidth. Reproduction of CompPath studies is further complicated by
the large number of proprietary and incompatible WSI file formats that
often impede data access and make it difficult to combine heterogeneous
data from different studies or sites. The Digital Imaging and
Communications in Medicine (DICOM) standard~\citep{bidgood1997} is an
internationally accepted standard for storage and communication of
medical images. It is universally used in radiology and other medical
disciplines, and has great potential to become the uniform standard for
pathology images as well~\citep{herrmann2018}. However, until now, there
have been few pathology data collections provided in DICOM format.

\subsection{NCI Imaging Data Commons}

The National Cancer Institute (NCI) Imaging Data Commons (IDC) is a new
cloud-based repository within the US national Cancer Research Data
Commons (CRDC)~\citep{fedorov2021}. A central goal of the IDC is to
improve the reproducibility of data-driven cancer imaging research. For
this purpose, the IDC provides large public cancer image collections
according to the FAIR principles.

Besides pathology images (brightfield and fluorescence) and their
metadata, the IDC includes radiology images (e.g., CT, MR, and PET)
together with associated image analysis results, image annotations, and
clinical data providing context about the images. At the time of writing
this article, the IDC contained 128 data collections with more than
63,000 cases and more than 38,000 WSI from different projects and sites.
The collections cover common tumor types, including carcinomas of the
breast, colon, kidney, lung, and prostate, as well as rarer cancers such
as sarcomas or lymphomas. Most of the WSI collections originate from The
Cancer Genome Atlas (TCGA)~\citep{tcga} and Clinical Proteomic Tumor
Analysis Consortium (CPTAC)~\citep{cptac} projects and were curated by
The Cancer Imaging Archive (TCIA)~\citep{clark2013}. These collections
are commonly used in the development of CompPath
methods~\citep{coudray2018, saltz2018, khosravi2018, noorbakhsh2020}.

The IDC implements the FAIR principles as follows:

Interoperability: While the original WSIs were provided in proprietary,
vendor-specific formats, the IDC harmonized the data and converted them
into the open, standard DICOM format~\citep{herrmann2018}. DICOM defines
data models and services for storage and communication of medical image
data and metadata, as well as attributes for different real-world
entities (e.g., patient, study) and controlled terminologies for their
values. In DICOM, a WSI corresponds to a ``series'' of DICOM image
objects that represent the digital slide at different resolutions. Image
metadata are stored as attributes directly within the DICOM objects.

Accessibility: The IDC is implemented on the Google Cloud Platform
(GCP), enabling cohort selection and analysis directly in the cloud.
Since IDC data are provided as part of the Google Public Datasets
Program, it can be freely accessed from cloud or local computing
environments. In the IDC, DICOM objects are stored as individual DICOM
files in Google Cloud Storage (GCS) buckets and can be retrieved using
open, free, and universally implementable tools.

Findability: Each DICOM file in the IDC has a persistent universally
unique identifier (UUID)~\citep{leach2005}. DICOM files in storage
buckets are referenced through GCS URLs, consisting of the bucket URL
and the UUID of the file. Images in the IDC are described with rich
metadata, including patient (e.g., age, sex), disease (e.g., subtype,
stage), study (e.g., therapy, outcome), and imaging-related data (e.g.,
specimen handling, scanning). All DICOM and non-DICOM metadata are
indexed in a BigQuery database~\citep{google2022bigquerydicom} that can
be queried programmatically using standard Structured Query Language
(SQL) statements (see section ``IDC data access''), allowing for an
exact and persistent definition of cohorts for subsequent analysis.

Reusability: All image collections are associated with detailed
provenance information but stripped of patient-identifiable information.
Most collections are released under data usage licenses that allow
unrestricted use in research studies.

\subsection{Objective}

This paper explores how the IDC and cloud ML services can be used in
combination for CompPath studies and how this can facilitate
reproducibility. This paper is also intended as an introduction to how
the IDC can be used for reproducible CompPath research. Therefore,
important aspects such as data access are described in more detail in
the Methods section.

\section{Methods}

\subsection{Overview}

We implemented two CompPath experiments using data collections from the
IDC and common ML services (Figure~1). Since the computing environments
provided by cloud ML services are all virtualized, two identically
configured instances may run different host hardware and software (e.g.,
system software versions, compiler settings)~\citep{gundersen2022}. To
investigate if and how this affects reproducibility, both experiments
were executed multiple times, each in a new instance of the respective
ML service.

The experiments are based on a basic CompPath analysis method that
addresses a use case representative of common CompPath
tasks~\citep{cruzroa2017, campanella2019, wang2020, iizuka2020, coudray2018}:
the automatic classification of entire WSI of hematoxylin and eosin
(H\&E)-stained lung tissue sections into either non-neoplastic (normal),
lung adenocarcinoma (LUAD), or lung squamous cell carcinoma (LSCC/LUSC).

Experiment~1 replays the entire development process of the method,
including model training and validation. Experiment~2 performs inference
with a trained model on independent data. The model trained in
Experiment~1 was used as the basis for Experiment~2. The two experiments
were conducted with different collections in the IDC:
TCGA-LUAD/LUSC~\citep{tcga-luad2016, tcga-lusc2016} and
CPTAC-LUAD/LSCC~\citep{cptac-luad2018, cptac-lscc2018}, respectively.
While both the TCGA and the CPTAC collections cover H\&E-stained lung
tissue sections of the three classes considered (Figure~2), they were
created by different clinical institutions using different slide
preparation techniques.

\begin{figure*}[h]
\centering
\includegraphics[width=\textwidth]{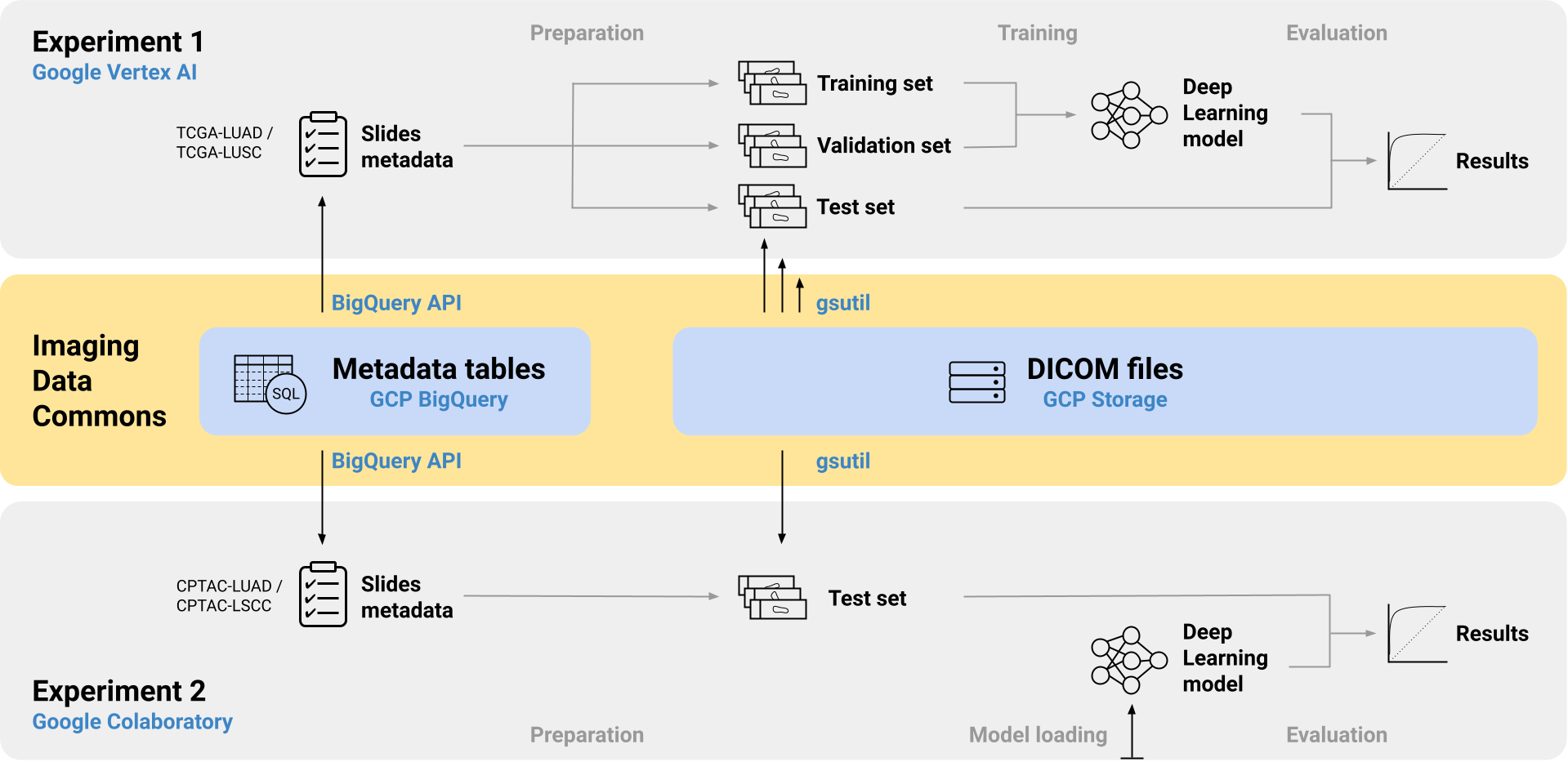}
\caption{Overview of the workflows of both experiments and
their interactions with the IDC.}
\end{figure*}

\begin{figure*}[h]
\centering
\includegraphics[width=\textwidth]{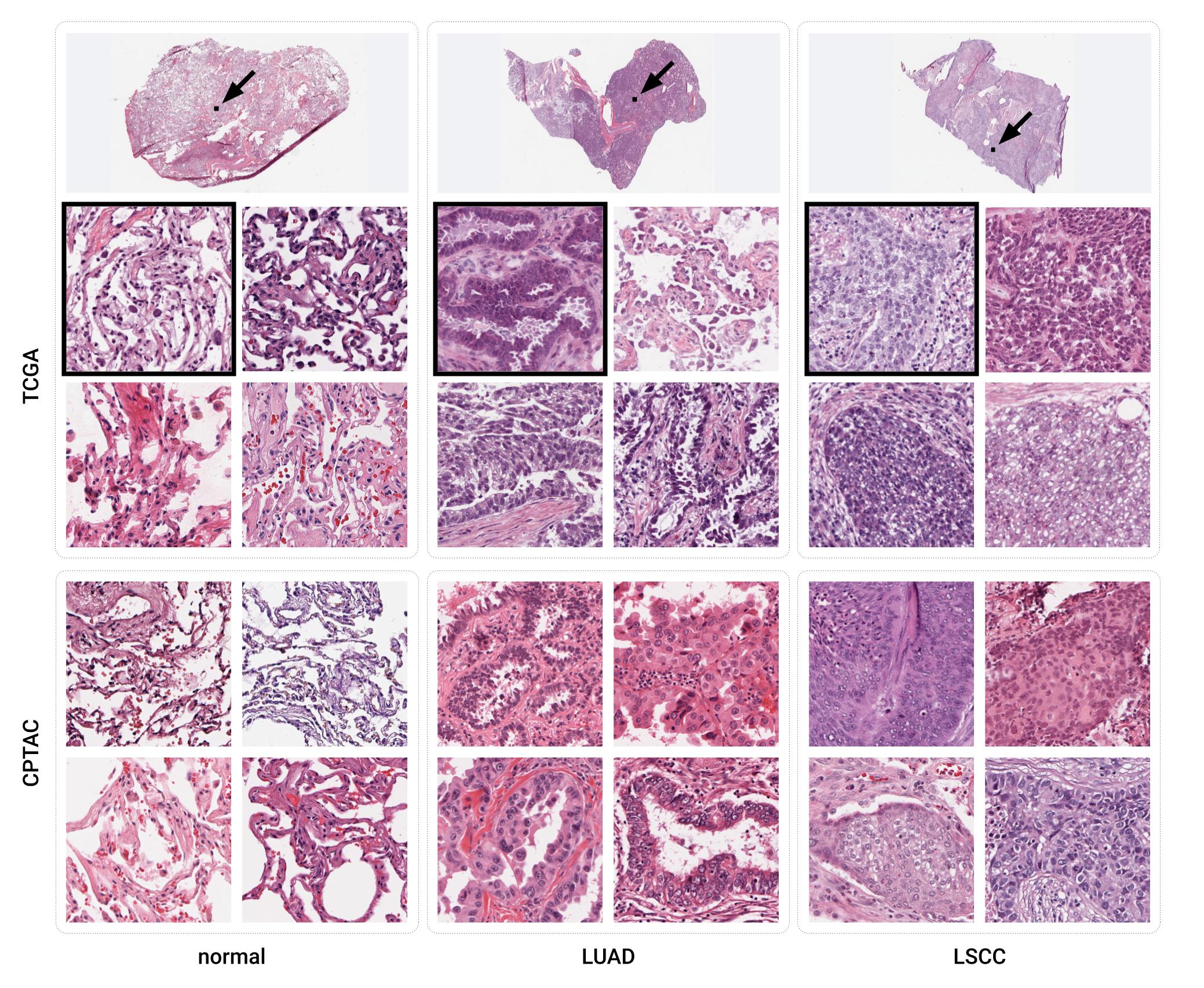}
\caption{Example tiles of the three classes considered from
the TCGA and CPTAC datasets. The width of each tile is 256~µm. The black
boxes marked with arrows in the whole slide images on top show the
boundaries of the upper left tiles of the TCGA data set.}
\end{figure*}

\subsection{Implementation}

Both experiments were implemented as standalone Jupyter notebooks that
are available open source~\citep{github_repo_experiments}. To enable
reproducibility, care was taken to make operations deterministic, e.g.,
by seeding pseudo-random operations, fixing initial weights for network
training, and by iterating over unordered container types in a defined
order. Utility functionality was designed as generic classes and
functions that can be reused for similar use cases.

As the analysis method itself is not the focus of this paper, we adopted
the algorithmic steps and evaluation design of a lung tumor
classification method described in a widely cited study by
Coudray~et~al.~\citep{coudray2018}. The method was chosen because it is
representative of common CompPath tasks and easy to understand. Our
implementation processed images at a lower resolution, which is
significantly less computationally expensive.

In our analysis workflow, a WSI was subdivided into non-overlapping
rectangular tiles, each measuring 256 pixels at a resolution of 1~µm/px.
Tiles containing less than 50\% tissue, as determined by pixel value
statistics, were discarded. Each tile was assigned class probabilities
by performing multi-class classification using an
InceptionV3~CNN~\citep{szegedy2015}. The per-tile results were finally
aggregated to a single classification of the entire slide. The workflow
is visualized in~Figure~3 and a detailed description is provided in the
respective notebooks.

\begin{figure*}[h]
\centering
\includegraphics[width=\textwidth]{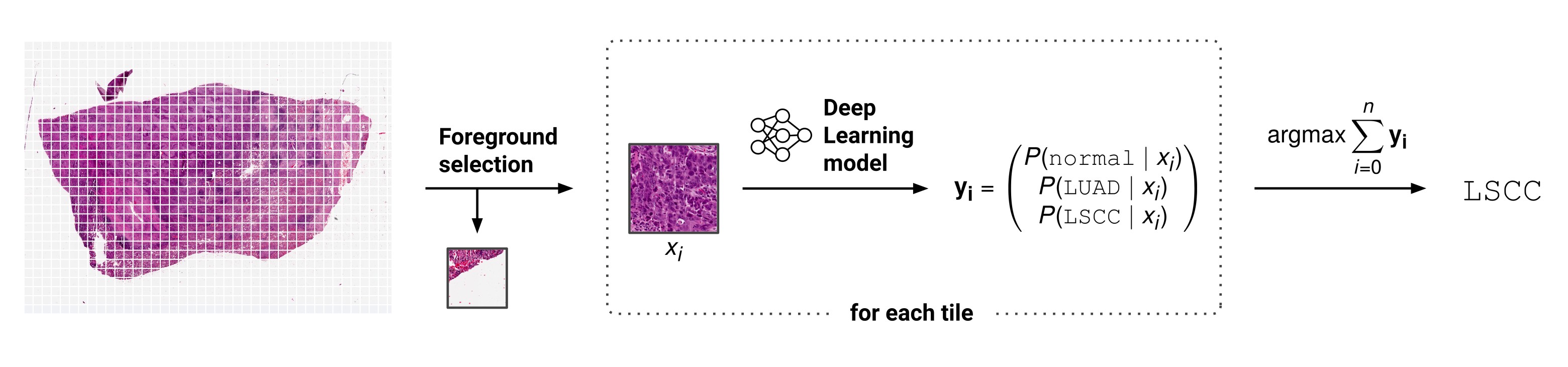}
\caption{Illustration of the CompPath analysis method. Slides
were subdivided into non-overlapping rectangular tiles discarding those
with more background than tissue. Each tile was assigned class
probabilities using a neural network performing multi-class
classification. Slide-based class values were determined by aggregating
the tile-based results.}
\end{figure*}

In Experiment~1, the considered slides were divided into training,
validation, and test sets with proportions of 70\%, 15\%, and 15\%,
respectively. To keep the sets independent and avoid overoptimistic
performance estimates~\citep{homeyer2022}, we ensured that slides from a
given patient were assigned to only one set, which resulted in 705, 151
and 153 patients per subset. The data collections used did not contain
annotations of tumor regions, but only one reference class value per
WSI. Following the procedure used by Coudray et al., all tiles were
considered to belong to the reference class of their respective slide.
Training was performed using a categorical cross-entropy loss between
the true class labels and the predicted class probabilities, and the
RMSProp optimizer with minimal adjustments to the default hyperparameter
values~\citep{keras2022rmsprop}. The epoch with the highest area under
the receiver operating characteristic (ROC) curve (AUC) on the
validation set was chosen for the final model.

\subsection{IDC data access}

For most CompPath studies, one of the first steps is to select relevant
slides using appropriate metadata. In the original data collections,
parts of the metadata were stored in the image files and other parts in
separate files of different formats (e.g., CSV, JSON files). In order to
select relevant slides, the image and metadata first had to be
downloaded in their entirety and then the metadata had to be processed
using custom tools. With the IDC, data selection can be done by
filtering a rich set of DICOM attributes with standard BigQuery SQL
statements (Figure~4). The results are tables in which rows represent
DICOM files and columns represent selected metadata attributes. As this
facilitates the accurate and reproducible definition of the data subsets
used in the analysis, these statements are described in more detail
below.

\begin{figure*}[h]
\centering
\includegraphics[width=.75\textwidth]{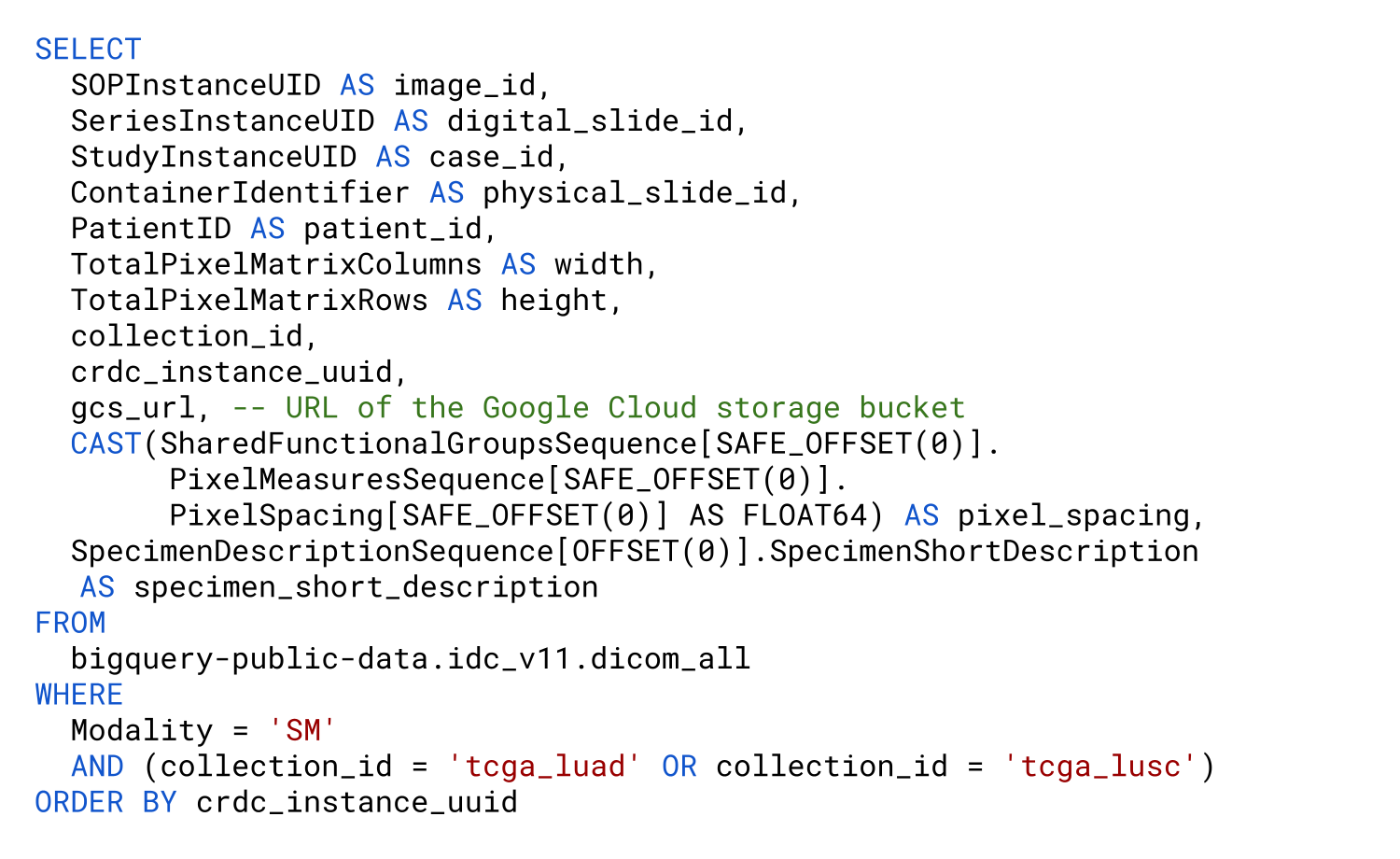}
\caption{Generic example of a BigQuery SQL statement for
compiling slide metadata. The result set is limited to slide microscopy
images, as indicated by the value ``SM'' of the DICOM attribute
``Modality'', from the collections ``TCGA-LUAD'' and ``TCGA-LUSC''.}
\end{figure*}

An SQL query for selecting WSI in the IDC generally consists of at least
a SELECT, a FROM and a WHERE clause. The SELECT clause specifies the
metadata attributes to be returned. The IDC provides a wealth of
metadata attributes, including image-, patient-, disease-, and
study-level properties. The attribute ``gcs\_url'' is usually selected
because it stores the GCS URL needed to access the DICOM file. The FROM
clause refers to a central table ``dicom\_all'' which summarizes all
DICOM attributes of all DICOM files. This table can be joined with other
tables containing additional project-specific metadata. Crucial to
reproducibility is that all IDC data are versioned: Each new release of
the IDC is represented as a new BigQuery dataset, keeping the metadata
for the previous release and the corresponding DICOM files accessible
even if they are modified in the new release. The version to use is
specified via the dataset specifier in fully qualified table names. All
experiments in this manuscript were conducted against IDC data version
11, i.e., the BigQuery table
``bigquery-public-data.idc\_v11.dicom\_all''. The WHERE clause defines
which DICOM files are returned by imposing constraints for certain
metadata attributes. To guarantee reproducibility, it is essential to
not use SQL statements that are non-deterministic (e.g., those that
utilize ANY\_VALUE) and conclude the statement with an ORDER BY clause,
which ensures that results are returned in a sorted order.

The two experiments considered in this paper also begin with the
execution of a BigQuery SQL statement to select appropriate slides and
required metadata from the IDC. A detailed description of the statements
is given in the respective notebooks. Experiment~1 queries specific
H\&E-stained tissue slides from the TCGA-LUAD/LUSC collections,
resulting in 2163 slides (591 normal, 819 LUAD, 753 LSCC). Experiment~2
uses a very similar statement to query the slides from the
CPTAC-LUAD/LSCC collections, resulting in 2086 slides (743 normal, 681
LUAD, 662 LSCC).

Once their GCS URLs are known, the selected DICOM files in the IDC can
be accessed efficiently using the open source tool
``gsutil''~\citep{google2022gsutil} or any other tool that supports the
Simple Storage Service (S3) API. During training in Experiment~1, image
tiles of different WSI had to be accessed repeatedly in random order. To
speed up this process, all considered slides were preprocessed and the
resulting tiles were extracted from the DICOM files and cached as
individual PNG files on disk before training. In contrast, simply
applying the ML method in Experiment~2 required only a single pass over
the tiles of each WSI in sequential order. Therefore, it was feasible to
access the respective DICOM files and iterate over individual tiles at
the time they were needed for the application of the ML method.

\subsection{Cloud ML services}

The two experiments were conducted with two different cloud ML services
of the GCP---Vertex AI and Google Colaboratory. Both services offer
virtual machines (VMs) preconfigured with common ML libraries and a
JupyterLab-like interface that allows editing and running notebooks from
the browser. They are both backed with extensive computing resources
including state-of-the-art GPUs or TPUs. The costs of both services
scale with the type and duration of use for the utilized compute and
storage resources. To use any of them with the IDC, a custom Google
Cloud project must be in place for secure authentication and billing, if
applicable.

Since training an ML model is much more computationally intensive than
performing inference, we conducted Experiment~1 with Vertex AI and
Experiment~2 with Google Colaboratory. Vertex AI can be attached to
efficient disks for storage of large amounts of input and output data,
making it more suitable for memory-intensive and long-running
experiments. Colaboratory, on the other hand, offers several less
expensive payment plans, with limitations in the provided computing
resources and guaranteed continuous usage times. Colaboratory can even
be used completely free of charge, with a significantly limited
guaranteed GPU usage time (12 hours at the time of writing). This makes
Colaboratory better suited for smaller experiments or exploratory
research.

\subsection{Evaluation}

Experiment~1 was performed using a common Vertex AI VM configuration (8
vCPU, 30 GB memory, NVIDIA T4 GPU, Tensorflow Enterprise 2.8
distribution). Experiment~2 was performed with Colaboratory runtimes
(2--8 vCPU, 12--30 GB memory). When using Google Colaboratory for
Experiment~2, we were able to choose between different GPU types,
including NVIDIA T4 and NVIDIA P100 GPUs. Since it has been suggested
that the particular type of GPU can affect
results~\citep{nagarajan2019}, all runs of Experiment~2 were repeated on
both GPUs, respectively. Runs with NVIDIA T4 were performed with the
free version of Colaboratory, while runs with NVIDIA P100 were performed
in combination with a paid GCE Marketplace VM, which was necessary for
guaranteed use of this GPU.

For each run of an experiment, classification accuracy was assessed in
terms of class-specific, one vs.~rest AUC values based on the
slide-level results. In addition, 95\% confidence intervals of the AUC
values were computed by 1000-fold bootstrapping over the slide-level
results.

To speed up Experiment~2, only a random subset of 300 of the selected
slides (100 normal, 100 LUAD, 100 LSCC) was considered in the analysis,
which was approximately the size of the test set in Experiment~1.

\section{Results}

The evaluation results of both experiments are summarized in Table~1. It
became apparent that none of the experiments was perfectly reproducible
and there were notable deviations in the results of repeated runs. In
Experiment~1, AUC values differed by up to 0.045 between runs. In
Experiment~2, there were also minimal deviations in the AUC values of
the different runs, but none of these were greater than 0.001. These
deviations occurred regardless of whether the runs were executed on the
same GPU type or not.

The classification accuracy of the method trained in Experiment~1
appears satisfactory when evaluated on the TCGA test set and comparable
to the results of a similar study based on the same TCGA
collections~\citep{coudray2018}. When applied to the CPTAC test set in
Experiment~2, the same model performed substantially worse (Figure~5).

Experiment~1 took an order of magnitude longer to complete (mean runtime
of 1~d~18~h ±1~h) than Experiment~2 (mean runtime of 1~h~54 min~±23~min
with NVIDIA T4 and mean runtime of 1~h~28 min~±8~min with NVIDIA P100).
The ML service usage charges for Experiment~1 were approximately US\$ 32
per run. With the free version of Colaboratory, Experiment~2 was
performed at no cost, while runs with the GCE Marketplace VM cost
approximately US\$ 2 per run.

\begin{table*}[h]
    \centering\small
    \begin{tabularx}{\textwidth}{lllllllll}
    \toprule
     ~ & ~ & ~ & \multicolumn{2}{c}{\textbf{normal}} & \multicolumn{2}{c}{\textbf{LUAD}} & \multicolumn{2}{c}{\textbf{LSCC}} \\ \cmidrule{4-9}
 \textbf{Experiment} & \textbf{ML Service (GPU)} & \textbf{Run} & \textbf{AUC} & \multicolumn{1}{c}{\textbf{CI}} & \textbf{AUC}  & \multicolumn{1}{c}{\textbf{CI}} & \textbf{AUC}  & \multicolumn{1}{c}{\textbf{CI}} \\ \midrule

		Experiment 1 & Vertex AI (T4) & 1 & 0.994 & [0.987, 0.999] & 0.942 & [0.914, 0.968] & 0.937 & [0.904, 0.964] \\  
		~ & ~ & 2 & 0.981 & [0.964, 0.994] & 0.898 & [0.860, 0.937] & 0.914 & [0.875, 0.946] \\ 
		~ & ~ & 3 & 0.992 & [0.983, 0.999] & 0.939 & [0.909, 0.964] & 0.918 & [0.881, 0.949] \\ 
		~ & ~ & 4 & 0.994 & [0.986, 0.999] & 0.928 & [0.895, 0.958] & 0.910 & [0.865, 0.947] \\
		~ & ~ & 5 & 0.989 & [0.979, 0.997] & 0.930 & [0.895, 0.959] & 0.892 & [0.838, 0.934] \\ \midrule
		Experiment 2 & Colaboratory (T4) & 1 & 0.811 & [0.746, 0.871] & 0.698 & [0.633, 0.759] & 0.850 & [0.802, 0.899] \\ 
        ~ & ~ & 2 & 0.811 & [0.746, 0.871] & 0.698 & [0.633, 0.759] & 0.850 & [0.802, 0.899]  \\ 
        ~ & ~ & 3 & 0.811 & [0.747, 0.870] & 0.698 & [0.636, 0.758] & 0.851 & [0.800, 0.896]  \\ 
        ~ & ~ & 4 & 0.811 & [0.748, 0.869] & 0.698 & [0.632, 0.758] & 0.851 & [0.802, 0.896]  \\ 
        ~ & ~ & 5 & 0.811 & [0.748, 0.872] & 0.698 & [0.627, 0.759] & 0.851 & [0.799, 0.896]  \\
        ~ & Colaboratory (P100) & 1 & 0.811 & [0.746, 0.874] & 0.698 & [0.630, 0.758] & 0.851 & [0.802, 0.896] \\ 
        ~ & ~ & 2 & 0.811 & [0.747, 0.873] & 0.698 & [0.627, 0.760] & 0.850 & [0.802, 0.897]  \\ 
        ~ & ~ & 3 & 0.811 & [0.747, 0.873] & 0.698 & [0.627, 0.760] & 0.850 & [0.802, 0.897]  \\ 
		~ & ~ & 4 & 0.811 & [0.747, 0.873] & 0.698 & [0.627, 0.760] & 0.850 & [0.802, 0.897]  \\ 
		~ & ~ & 5 & 0.811 & [0.747, 0.873] & 0.698 & [0.627, 0.760] & 0.850 & [0.802, 0.897]  \\
	\bottomrule
    \end{tabularx}
    \caption{Class-specific, slide-based AUC values and 95\% confidence intervals (CI) obtained through multiple runs of both experiments.}
	\label{table:experiment_results}
\end{table*}

\begin{figure*}[h]
\centering
\includegraphics[width=\textwidth]{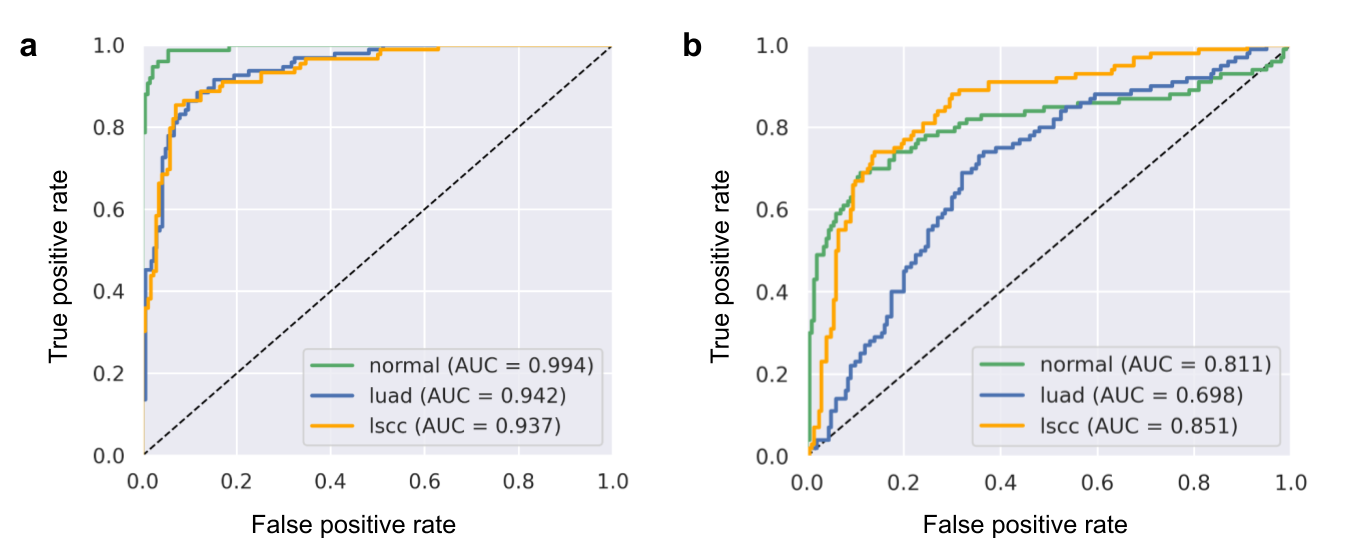}
\caption{One-vs-rest ROC curves for the multi-class classification as obtained in (a) the first run of Experiment~1 using Vertex AI and (b) the first run of Experiment~2 using Colaboratory (T4).}
\end{figure*}

\section{Discussion}

The aim of this study was to investigate how CompPath studies can be
made reproducible through the use of cloud-based computing environments
and the IDC as the source of input data. Although the same code was run
with the same data using the same ML services and care was taken that
operations were deterministic (see section ``Implementation''), we
observed small deviations in the results of repeated runs. We did not
investigate whether the deviations originate from differences in the
hardware and software used by the hosts of the virtual computing
environments, or whether they are due to randomness resulting from
parallel processing~\citep{gundersen2022}. The greater variability in
the results of Experiment~1 can possibly be explained by its higher
computational complexity. Although the observed deviations appear
negligible for many applications, they represent a practical upper limit
for reproducibility. Such issues are likely to occur in any computing
environment. As outlined below, we argue that the IDC can help to
approach this reproducibility limit.

We chose Jupyter notebooks and cloud ML services to address the first
two reproducibility challenges mentioned in the Introduction: specifying
the analysis method and setting up the computing environment. With the
IDC, we were able to tackle the third reproducibility challenge with
respect to the special requirements of CompPath: specifying and
accessing the data.

By providing imaging data collections according to the FAIR principles,
the IDC facilitates precise definition of the datasets used in the
analysis and ensures that the exact same data can be reused in follow-up
studies. Since metadata on acquisition and processing can be included as
DICOM attributes alongside the pixel data, the risk of data confusion
can be greatly reduced. The IDC also facilitated the use of cloud ML
services because it makes terabytes of WSI data efficiently accessible
by on-demand compute resources. We consider our experiments to be
representative of common CompPath applications. Therefore, the IDC
should be similarly usable for other CompPath studies.

The results of Experiment~2 also reveal the transferability of the model
trained in Experiment~1 to independent data. Although the majority of
slides were correctly classified, AUC values were significantly lower,
indicating that the model is only transferable to a limited extent and
additional training is needed. Since all IDC data collections (both the
image pixel data and the associated metadata) are harmonized into a
standardized DICOM representation, testing transferability to a
different dataset required only minor adjustments to our BigQuery SQL
statement. In the same way, the IDC makes it straightforward to use
multiple datasets in one experiment or to transfer an experimental
design to other applications.

\subsection{Limitations}

Using cloud ML services comes with certain trade-offs. Conducting
computationally intensive experiments requires setting up a payment
account and paying a fee based on the type and duration of the computing
resources used. Furthermore, although the ML services are widely used
and likely to be supported for at least the next few years, there is no
guarantee that they will be supported in the long term and support the
specific configuration of the computing environment used (e.g., software
version, libraries). Those who do not want to make these compromises can
also access IDC data collections without using ML services, both in the
cloud and on-premises. Even if this means losing the previously
mentioned advantages with regard to the first two reproducibility
challenges, the IDC can still help to specify the data used in a clear
and reproducible manner.

Independent of the implementation, a major obstacle to the
reproducibility of CompPath methods remains their high computational
cost. A full training run often takes several days, making reproduction
by other scientists tedious. Performing model inference is generally
faster and less resource intensive when compared to model training.
Therefore, Experiment~2 runs well even with the free version of Google
Colaboratory, enabling others to reproduce it without spending money.
The notebook also provides a demo mode, which completes in a few
minutes, so anyone can easily experiment with applying the inference
workflow to arbitrary images from IDC.

At the moment, the IDC exclusively hosts public data collections. New
data must undergo rigorous curation to de-identify (done by TCIA or data
submitter) and harmonize images into standard representation (done by
IDC), which can require a significant effort. Therefore, only data
collections that are of general relevance and high quality are included
in the IDC. As a result, the data in the IDC were usually acquired for
other purposes than a particular CompPath application and cannot be
guaranteed to be representative and free of bias~\citep{varoquaux2022}.
Compiling truly representative CompPath datasets is very
challenging~\citep{homeyer2022}. Nevertheless, the data collections in
the IDC can provide a reasonable basis for exploring and prototyping
CompPath methods.

\subsection{Outlook}

The IDC is under continuous development and its technical basis is
constantly being refined, e.g., to support new data types or to
facilitate data selection and access. Currently, DICOM files in the IDC
can only be accessed as a whole from their respective storage buckets.
This introduces unnecessary overhead when only certain regions of a
slide need to be processed, and it may make it necessary to temporarily
cache slides to efficiently access multiple image regions (see section
``IDC data access''). Future work should therefore aim to provide
efficient random access to individual regions within a WSI. For maximum
portability, such access should ideally be possible via standard DICOM
network protocols such as DICOMweb~\citep{dicomweb2022, herrmann2018}.

The IDC is continuously being expanded to support even more diverse
CompPath applications. For instance, images collected by the Human Tumor
Atlas Network (HTAN) that provide rich, multispectral information on
subcellular processes~\citep{rozenblattrosen2020} have recently been
added. The IDC is integrated with other components of the CRDC, such as
the Genomic Data Commons~\citep{grossman2016} or the Proteomic Data
Commons~\citep{pdc2022}. This opens up many more potential CompPath
applications involving tissue images and different types of molecular
cancer data~\citep{schneider2022}.

\subsection{Conclusion}

We demonstrated how the IDC can facilitate the reproducibility of
CompPath studies. Implementing future studies in a similar way can help
other researchers and peer reviewers to understand, validate and advance
the analysis approach.

\section{Author Contributions}

DPS and AH conceived and carried out the study. AH and AF supervised the
project. AF, MDH, DAC, HH, WC, WJRL, SP and RK supported the study in
different ways, e.g., by providing data, supporting set-up of the
computing infrastructure, interpretation of the results and giving
general advice. AH and DPS drafted the manuscript. All authors
critically revised the manuscript and expressed their consent to the
final version.

\section{Declaration of Competing Interest}

The authors declare no conflicts of interest.

\section{Acknowledgements}

The authors thank Lars Ole Schwen for advice on deterministic
implementations of machine learning algorithms and Tim-Rasmus Kiehl for
advice on tissue morphology.

The results published here are in whole or part based upon data
generated by the TCGA Research Network and the National Cancer Institute
Clinical Proteomic Tumor Analysis Consortium (CPTAC).

This project has been funded in whole or in part with Federal funds from
the National Cancer Institute, National Institutes of Health, under Task
Order No.~HHSN26110071 under Contract No.~HHSN261201500003l.

\bibliography{ms}

\end{document}